\documentclass[10pt,twocolumn,letterpaper]{article}

\usepackage[letterpaper,margin=1in]{geometry}
\usepackage{times}
\usepackage{graphicx}
\usepackage{amsmath,amssymb}
\usepackage{booktabs}
\usepackage{algorithm}
\usepackage{algpseudocode}
\usepackage[font=small]{caption}
\usepackage{microtype}
\usepackage[numbers,sort&compress]{natbib}
\usepackage{xcolor}
\definecolor{linkblue}{rgb}{0.21,0.49,0.74}
\usepackage[breaklinks,colorlinks,allcolors=linkblue]{hyperref}

\newcommand{\forth}{\textsc{forth}}

\title{\forth: Grounding Tactile VLA Execution via Forward-Only Proxy-Control Pairing}
\author{Junde Li\textsuperscript{1}, Wei-Han Yu\textsuperscript{1}, Xiangyi Chu\textsuperscript{1}, Yaoxin Li\textsuperscript{1}, Jiawei Xu\textsuperscript{1}, Ka-Fai Un\textsuperscript{1}, Pui-In Mak\textsuperscript{1}\\[4pt]
\textsuperscript{1}University of Macau}
\date{}
\hypersetup{pdftitle={FORTH: Grounding Tactile VLA Execution via Forward-Only Proxy-Control Pairing},
            pdfauthor={Junde Li, Wei-Han Yu, Xiangyi Chu, Yaoxin Li, Jiawei Xu, Ka-Fai Un, Pui-In Mak},
            pdfsubject={Preprint}}

\begin{document}
\maketitle

\setlength{\abovedisplayskip}{3pt plus 1pt minus 1pt}
\setlength{\belowdisplayskip}{3pt plus 1pt minus 1pt}
\setlength{\abovedisplayshortskip}{1pt plus 1pt}
\setlength{\belowdisplayshortskip}{1pt plus 1pt}

\begin{abstract}
Vision-Language-Action (VLA) policies are widely credited with force control, operating on the unverified assumption that proxy loss minimization implies physical force grounding. In this work, we demonstrate that this assumption breaks down at three core deployment levels: frozen visual backbones discard contact geometry, downstream losses mask token-level gradient alignment, and discrete indicators trigger optimization collapse. To address these disconnections, we present \forth{}, a forward-only measurement harness and execution framework that explicitly pairs the optimization objective at each level with its physical control observable. \forth{} incorporates three grounded pillars: a layer-local, forward-only tactile encoder that preserves spatial contact geometry without backpropagation, a token-parallel zeroth-order adapter enabling activation-space updates, and \textit{SlipGuard}, a continuous tactile verification gate that fires on incipient slip. On an elasto-plastic structural shell, the \textit{SlipGuard} controller peaks at $29\times$ lower grip force than the reflex ramp of the same shell, preventing structural crush where the baseline leaves permanent deformation. Furthermore, \forth{} matches state-of-the-art execution performance with a $97.6\%$ average success rate across three evaluated LIBERO suites, while reducing peak activation memory by $7.12\times$ compared to a backpropagated step on the same backbone. Enhanced perception enables deeper physical intelligence. The code and models are open-sourced at \url{https://github.com/Jordanlllll/FORTH}.
\end{abstract}

\begin{figure*}[t]
\centering
\includegraphics[width=\textwidth]{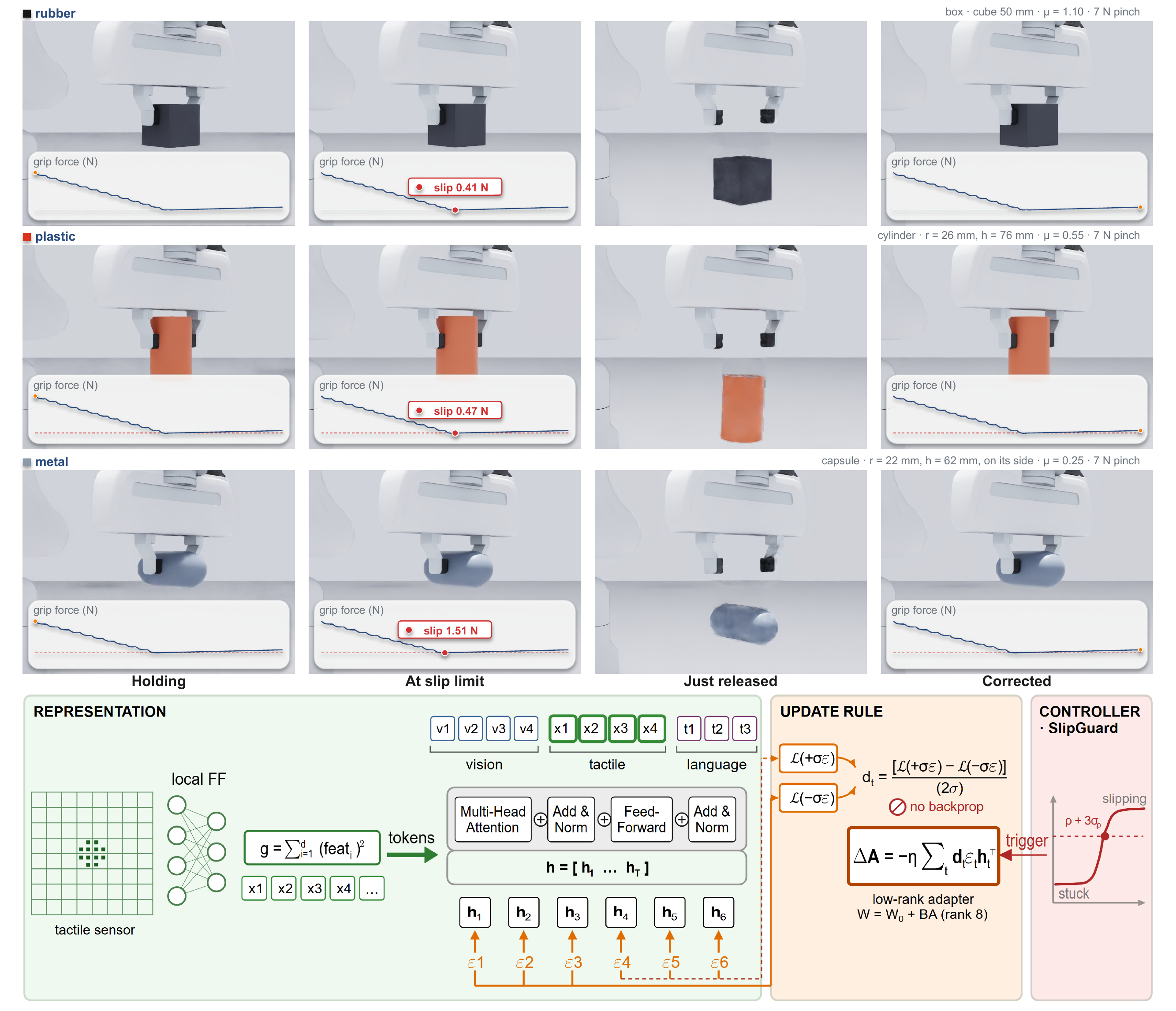}
\caption{\textbf{\forth{} architecture and experimental testbed.}
\textbf{Top:} Slip detection and recovery across three material regimes (rubber, plastic, and metal) driven past their
friction limits. While uncorrected execution leads to object dropping or crushing at the slip limit, \textit{SlipGuard}
dynamically restores stable grasp force.
\textbf{Bottom:} The three execution levels of \forth{}: (1) \textit{Representation}: a local Forward-Forward (FF)
tactile trunk; (2) \textit{Update Rule}: a token-parallel zeroth-order low-rank adapter on the action head operating
without backpropagation; and (3) \textit{Controller}: \textit{SlipGuard}, a shear-ratio gate triggering two-frame debounced corrective
adjustments.}
\label{fig:frames}
\end{figure*}

\section{Introduction}\label{sec:intro}

Vision-Language-Action (VLA) policies are rapidly advancing from tabletop picking to contact-rich manipulation. To
operate effectively under contact, state-of-the-art systems increasingly augment pretrained visual backbones with
tactile or force sensing modalities
\cite{li2026atvla,li2026forcevla2,feng2026tactilereflex,zhang2026retouch}. Concurrently, tactile
sensing hardware and representations have matured: optical tactile sensors, high-fidelity contact simulators, and
learned representations provide policies with spatially resolved contact maps rather than scalar load readings
\cite{lambeta2020digit,si2021taxim,wang2020tacto,george2024vital,xiang2026tactilesurvey}. While an emerging family of
tactile VLAs reports force-aware execution across tasks like insertion, wiping, and grasping
\cite{cheng2025omnivtla,ayral2026reactive,chen2026ptld,mao2026vibeact,zhou2026touchworld,cheng2026learning,yuan2026dextacwam,zhang2026restacvla,min2026geometric,meng2026reach,krohn2026multires,zhang2026detac},
performance across these systems is predominantly evaluated using a single proxy metric per task level. Such scalar
metrics obscure the underlying physical reality: the critical quantities governing whether an object survives an
episode are rarely measured or reported together. This discrepancy stems from a missing evaluation and reporting convention.

Existing research programs address isolated compute and memory bottlenecks without closing this measurement gap.
Layer-local learning objectives replace global backpropagation with forward-only computations
\cite{hinton2022,li2024ffchip,ma2025ffint8,rub2025sparseff,jaiswal2026binaryff,nokland2019localerrorsignals,ren2022forwardgradient,krutsylo2025scalableff,gandhi2023extendingff,ororbia2023predictiveff,ahamed2023ffcontrastive},
yet report performance strictly via local loss values. Zeroth-order (ZO) and gradient-free optimizers trade
memory-intensive backward passes for forward queries
\cite{ars,mezo,eggroll,vlazo,zoadamm,gautam2024vrzofinetune,zhang2025zoflatminima,shang2025zoquantized,yang2025sharpzo},
evaluating only estimated gradient directional alignment. Parameter-efficient fine-tuning (PEFT) and quantization
constrain these updates within frozen backbones \cite{hu2021lora,dettmers2023qlora,houlsby2019adapters}, while
low-level controllers treat physical slip as a binary classification target scored by detection accuracy. Crucially,
an improvement in these algorithmic proxies can decouple from physical interaction quality---a lower local loss does
not inherently prevent object damage---and no current benchmark protocol
\cite{liu2023libero,mees2021calvin,james2019rlbench,mandlekar2021robomimic,zhu2020robosuite,makoviychuk2021isaacgym,nvidia2025isaaclab,chi2023diffusionpolicy,octo2024,kim2025openvlaoft,xiang2026contactbench,higuera2024sparsh,tsai2023realitygap}
systematically measures and reports these paired dynamics.

To resolve this decoupling, we introduce \forth{}, a unified harness governed by a paired metric convention across all
three execution levels: report the optimized algorithmic quantity alongside the physical quantity it aims to control,
accepting a policy level only when both metrics move in lockstep. Operating entirely forward-only and layer-locally on
frozen 4-bit backbones without global backward passes, \forth{} integrates three key components: (1) a tactile feature trunk
composed of two Forward-Forward (FF) blocks; (2) a token-parallel zeroth-order adapter of
rank 8 attached to the action head; and (3) \textit{SlipGuard}, a one-class shear-ratio gate with a bounded real-time
corrector. Rather than introducing a massive new backbone, \forth{} makes tactile policy execution measurable and
correctable directly on edge deployment hardware, requiring only 16 forward passes per update ($T{=}8$ tokens) instead of full
backpropagation.

This design yields a deployment-ready harness that acts dynamically before manipulation failure occurs. Empirically,
the forward-only tactile trunk matches backpropagated representation accuracy; the zeroth-order updater fine-tunes the
rank-8 adapter at a fraction of backward memory requirements; and \textit{SlipGuard} converts incipient
slip into a recoverable interaction across three slip severity regimes, where standard hand-tuned force ramps result
in permanent object crush. Figure~\ref{fig:frames} illustrates our testbed for recording physical contact dynamics
across three material bodies driven past their friction limits.

\section{Related Work}
\label{sec:related_work}

\subsection{Vision-Language-Action Policies}\label{sec:vla_policies}
\label{sec:rel-vla}

Vision-language-action models predict an action sequence from an image and an
instruction~\cite{kim2024openvla,black2024pi0,brohan2023rt2}. Their scale comes from a pretrained
multimodal backbone rather than from robot data, and the head that turns features into commands is
small enough to adapt on one GPU. Alongside them sit action-chunking transformers for low-cost bimanual
teleoperation~\cite{zhao2023actaloha}, tokenizers that compress a chunk to a short
sequence~\cite{pertsch2025fast} and open-world variants of the recipe~\cite{physicalintelligence2025,yang2026mantis,huang2026fastthinkact}.

Touch enters that stack in three ways: as semantic-aligned tokens projected into the same space as
text~\cite{cheng2025omnivtla}, as a residual representation added to a frozen visual
policy~\cite{zhang2026restacvla}, or as a predictive model that forecasts contact before it
occurs~\cite{zhou2026touchworld,yuan2026dextacwam,jing2026cand}. Neighboring designs fuse heterogeneous tactile
sensors behind one policy~\cite{krohn2026multires,li2026atvla}, learn the touch representation from future visual
supervision~\cite{cheng2026learning}, or name material and object properties from touch without closing
a control loop~\cite{yu2025demonstrating}.

A separate line learns force implicitly rather than conditioning on it~\cite{li2026forcevla2,gillman2026goalforce}: grip strength emerges from the task reward~\cite{openai2018dactyl,rubik2019,chen2026ptld}, and the force in the logs is whatever the policy found sufficient.

\subsection{Forward-Only Update Rules}\label{sec:forward_only}
\label{sec:rel-forward}

Forward-only learning replaces the backward pass with a local objective. Forward-Forward trains each
layer on its own contrastive goodness~\cite{hinton2022}, and the similarity kernel of layer
representations contracts toward rank one with depth~\cite{priceoflocality}. Local error signals and
forward gradients are the neighboring non-contrastive route~\cite{nokland2019localerrorsignals,ren2022forwardgradient};
contrastive, predictive-coding and scalable variants extend
it~\cite{ahamed2023ffcontrastive,ororbia2023predictiveff,gandhi2023extendingff,krutsylo2025scalableff}.
On hardware, INT8, sparse, binary-stochastic and closed-box trainers report lower energy, memory or
compute than backpropagation at the edge~\cite{ma2025ffint8,rub2025sparseff,jaiswal2026binaryff,li2024ffchip,spyra2025}.

Zeroth-order optimization estimates a descent direction from function evaluations alone, from the
black-box attack estimator~\cite{chen2017zoo} and random search~\cite{ars,eggroll} through adaptive
moments~\cite{zoadamm} to MeZO~\cite{mezo}, with variance-reduced variants~\cite{gautam2024vrzofinetune}
and convergence analyses~\cite{zhang2025zoflatminima}. Applications reach VLA
adaptation~\cite{vlazo}, quantized fine-tuning~\cite{shang2025zoquantized} and prompt
tuning~\cite{yang2025sharpzo}.

These methods are judged by downstream loss or recall. A population argument, in which independent per-token perturbations average out, predicts that estimator quality improves with token count; that prediction has not been measured against the true gradient on a physical signal.

Adaptation and evaluation practice supply the rest: low-rank and quantized
adapters~\cite{hu2021lora,dettmers2023qlora,houlsby2019adapters}, one-bit backbones~\cite{wang2025bitvla},
benchmarks and simulators~\cite{liu2023libero,mees2021calvin,james2019rlbench,xiang2026contactbench,makoviychuk2021isaacgym,nvidia2025isaaclab,mu2025robotwin},
and replication practice~\cite{norton2024mirrer,vaughan2012publishcode,luo2026toward}.

\subsection{Slip Detection and Tactile Sensing}\label{sec:tactile_sensing}
\label{sec:rel-tactile}

Slip detection has a reflex-control literature that predates learned policies. Visuo-tactile
classifiers run on multi-scale temporal convolutions~\cite{gao2023visuo}, and incipient-slip
measurement tracks the transition during in-hand pivoting~\cite{li2023incipient,romano2011grasp,boonstra2024incipient,sui2024incipient}. Geometric
reconstruction recovers extrinsic contact from touch and proprioception~\cite{min2026geometric}, and
the same contact signal drives sub-millimeter insertion~\cite{meng2026reach}.

A second class puts that signal in a fast loop under a slow planner: a reactive dexterous
controller~\cite{niu2026rex,ayral2026reactive}, or a vibration signal routed straight to
action~\cite{mao2026vibeact}.

On the representation side the split is between an online-refined tactile prediction that absorbs
contact drift~\cite{zhang2026retouch} and a decomposition that separates force from surface
texture~\cite{zhang2026detac}. Both matter for the representation level: an aggregate can be predictive of
contact and still carry no usable force.

Evaluation in this line is detection rate, latency and false positives on a held-out session, which scores the detector while the force it commands can stay wrong.

The substrate matured alongside the policies: gel-sensor hardware~\cite{lambeta2020digit},
simulators~\cite{si2021taxim,wang2020tacto}, visuo-tactile pretraining~\cite{george2024vital},
sim-to-real transfer~\cite{tsai2023realitygap} and a recent survey~\cite{xiang2026tactilesurvey}.

\subsection{Proxy-Based Evaluation}\label{sec:proxy_evaluation}

Reporting conventions here are inherited from benchmark suites that score task success alone, or at most one scalar
force reading beside it
\cite{liu2023libero,mees2021calvin,james2019rlbench,mandlekar2021robomimic,zhu2020robosuite,chi2023diffusionpolicy,octo2024,kim2025openvlaoft,xiang2026contactbench,higuera2024sparsh,fei2026liberoplus,engelbracht2026hoi}.
Adapter and parameter-efficient practice keeps the same convention: the reported columns are loss and success, and the
controlled physical quantity is not among them \cite{hu2021lora,dettmers2023qlora,houlsby2019adapters}. Contact-aware
instrumentation measures the sensor rather than the policy's outcome
\cite{lambeta2020digit,si2021taxim,wang2020tacto,george2024vital,tsai2023realitygap}. The pairing defined next
therefore keeps each line's objective as given and adds the column none of them reports.

\section{Proxy-Control Framework}\label{sec:framework}

\subsection{Deployment Stack Taxonomy}\label{sec:taxonomy}

The FORTH framework organizes the tactile Vision-Language-Action (VLA) pipeline into a three-level hierarchy, where each level is governed by a specific proxy signal that approximates a critical physical control quantity.

\subsection{Paired Evaluation Protocol}\label{sec:protocol}

The validity of these proxies is established through a paired evaluation protocol that runs four checks in cost order---floor, readout, validity, then mechanism, whose protocol labels are $1$, $3$, $4$, $2$.

Low-overhead checks run before expensive simulations or hardware trials, so evaluation cost accumulates in non-decreasing order:
\begin{equation}
C_{total} = \sum_{i=1}^{4} c_i, \quad c_i \le c_{i+1},
\label{eq:cost}
\end{equation}
where $C_{total}$ is the total cost, $c_i$ the cost of the $i$-th check and $c_{i+1}$ the cost of the next one. This structure allows FORTH to maintain high execution fidelity while reducing the need for exhaustive real-world testing.

\begin{figure*}[t]
\centering
\includegraphics[width=\textwidth]{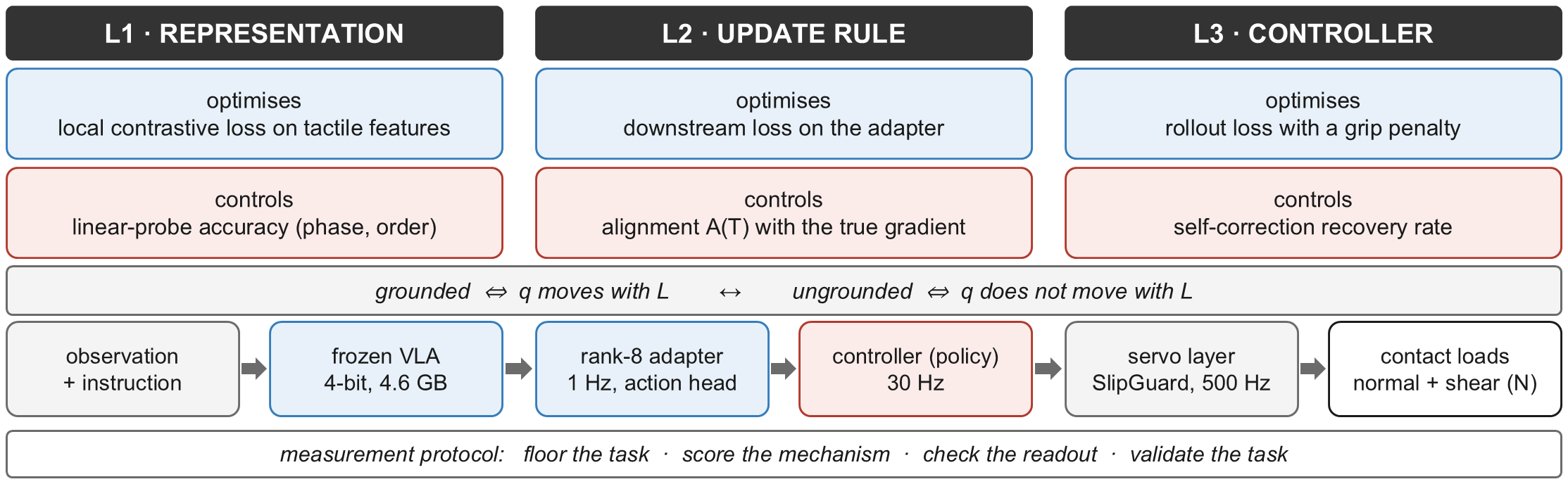}
\caption{\textbf{The FORTH evaluation harness and physical grounding protocol.} \forth{} pairs each level's proxy loss ($\mathbf{L1}$--$\mathbf{L3}$, one per level) with its physical control observable. \emph{Top}: grounded iff the observable $q$ moves monotonically with objective minimization. \emph{Middle}: observation to $500$\,Hz contact-load modulation at the tactile servo, with a $1$\,Hz rank-8 adapter and a $30$\,Hz policy on the 4-bit VLA. \emph{Bottom}: the four-step diagnostic protocol enforcing task validity before feature auditing.}
\label{fig:system}
\end{figure*}

Figure~\ref{fig:system} draws the three levels and names the stage each acts on (backbone, adapter, $30$\,Hz loop), The four checks run in cost order, so a task that fails a cheap check never reaches an expensive one.
\begin{equation}
\begin{split}
M(\theta_1, \theta_2) = {} & \mathbb{I}\bigl[\,|J(\theta_1) - J(\theta_2)| < \epsilon\,\bigr] \\
& \cdot \lVert q(\theta_1) - q(\theta_2) \rVert_2,
\end{split}
\label{eq:mismatch}
\end{equation}
where $M$ is the mismatch metric, $\theta_1$ and $\theta_2$ are two distinct sets of policy parameters, $\mathbb{I}$ is the indicator function, $J$ is the scored objective function, $\epsilon$ is a small scalar threshold, $q$ is the controlled physical quantity, and $\lVert\cdot\rVert_2$ denotes the Euclidean norm.
\begin{equation}
\text{sgn}(\Delta J(\theta, \delta)) = \text{sgn}(\Delta Q(\theta, \delta)),
\label{eq:sign}
\end{equation}
where $\text{sgn}$ is the signum function, $\Delta J$ is the change in the scored signal, $\theta$ represents the policy parameters, $\delta$ is a perturbation applied to those parameters, and $\Delta Q$ is the resulting change in the controlled physical quantity.

The protocol operationalizes the mismatch in Eq.~(\ref{eq:mismatch}) as the sign-agreement test of Eq.~(\ref{eq:sign}) and reports whether each level clears its rank-agreement threshold $\tau_{\text{rank}}$. Because this test is level-agnostic, the next three sections apply it in stack order, starting with the policy's input representation.

\section{Representation}\label{sec:representation}

\subsection{Layer-Local Trunk Architecture}\label{sec:layer_local_trunk}
The tactile representation processes local spatial interactions through a frozen layer-local trunk constructed from two Forward-Forward blocks, Eq.~(\ref{eq:patch})~\cite{hinton2022}. Let $\mathbf{x} \in \mathbb{R}^{48 \times d}$ denote the stacked sequence of projected patch embeddings:
\begin{equation}
\mathbf{x} = \mathrm{LinearProjection}\left(\mathrm{Patchify}(\mathbf{I})\right),
\end{equation}
where $\mathbf{x}$ is the sequence of patch tokens, $\mathrm{LinearProjection}$ maps raw visual-tactile interface patches into the latent vector space of dimension $d=128$, $\mathrm{Patchify}$ decomposes the continuous sensory surface tensor into discrete non-overlapping regions, and $\mathbf{I}$ represents the spatial surface sensory tensor extracted from the sensor interface.

Each block in the layer-local trunk updates its representations using a local goodness objective operating on scalar activations rather than propagating gradients back through the global backbone architecture. A given layer processes the incoming vector state to generate a zero-centered feature field, evaluated via a sum-of-squares goodness metric:
\begin{equation}
\mathbf{h} = \tanh(\mathbf{x}\mathbf{W}), \quad \mathbf{f} = \mathbf{h} - \boldsymbol{\mu}, \quad g = \sum_{i=1}^{d} (f_i)^2,
\label{eq:patch}
\end{equation}
where $\mathbf{h}$ is the activated hidden state, $\mathbf{x}$ the patch-token input, $\mathbf{W}$ the layer weight matrix, $\boldsymbol{\mu}$ the per-sample mean of $\mathbf{h}$ over the hidden dimension, $\mathbf{f}$ the centered feature, $d$ the feature dimension and $g$ the summed goodness of the sample.

Training proceeds locally within each block by contrasting positive sensory executions against negative counterfactual sensor states. A layer-specific scalar threshold $\theta \in \mathbb{R}$, a free parameter trained with the local objective and fixed at $2.0$ in the reported protocol, dictates the decision boundary separating valid tactile contacts from noise.

The optimization objective maximizes the scalar goodness for real physical interactions while suppressing goodness for synthetic or ungrounded states, using a localized logistic loss:
\begin{equation}
\mathcal{L}_{\mathrm{local}} = -\log \sigma(g - \theta) - \log \sigma(\theta - g_{\mathrm{neg}}),
\end{equation}
where $\mathcal{L}_{\mathrm{local}}$ is the layer-local forward-forward objective loss, $\log$ is the natural logarithm function, $\sigma$ is the standard logistic sigmoid function, $g$ is the scalar goodness computed on positive real tactile data, $\theta$ is the local layer threshold, and $g_{\mathrm{neg}}$ is the scalar goodness evaluated on negative counterfactual state samples.

Because only scalar goodness values cross block boundaries, the layer-local trunk maintains functional parity with a standard backpropagated network while avoiding end-to-end backpropagation. On a random validation split, this local trunk architecture achieves a tight trunk parity of $0.0004$ on the contact-phase task and $0.012$ on the temporal-order task.

\subsection{Force Grounding in Representation Features}\label{sec:force_grounding}
Unmodified frozen Vision-Language-Action backbones exhibit a fundamental failure mode termed force misgrounding. However, the ungrounded visual feature space is invariant to mechanical load, so the prompt changes the attention state but not the contact geometry, and the measured force does not move across command variations while the logit variance stays high.

To detect and evaluate this misgrounding diagnostic, a linear probe reads out contact presence directly from the layer-local trunk goodness metric:
\begin{equation}
\hat{c} = \mathbf{w}_{\mathrm{probe}}^{\top} \mathbf{f} + b_{\mathrm{probe}},
\end{equation}
where $\hat{c}$ is the predicted scalar contact variable, $\mathbf{f}$ the centered feature vector, $\mathbf{w}_{\mathrm{probe}}$ the probe weights and $b_{\mathrm{probe}}$ the probe bias.

The parity gap measures the functional discrepancy between representations learned by the forward-only local trunk and a globally backpropagated control trunk evaluated on the probing task:
\begin{equation}
\Delta_{\mathrm{parity}} = \left| \mathrm{acc}_{\mathrm{local}} - \mathrm{acc}_{\mathrm{backprop}} \right|,
\end{equation}
where $\Delta_{\mathrm{parity}}$ is the absolute accuracy gap, $\mathrm{acc}_{\mathrm{local}}$ the probe accuracy of the layer-local trunk and $\mathrm{acc}_{\mathrm{backprop}}$ that of a backpropagated trunk on the same split.

Across all evaluated samples, the minimal parity gap confirms that the layer-local trunk preserves critical contact representations without modifying the frozen vision-language backbone weights.

\subsection{Force Channels in Policy Input}\label{sec:force_channels}
Physical force signals cannot be introduced into the VLA execution policy as instantaneous scalar readings. To make the tactile signal usable for robust proxy control, force must enter the network as a temporally aggregated feature channel structured alongside spatial patch representations.

We test this interface with matched arms and controls. The channel integrates force across a sliding temporal window before policy decoding, which decouples language-level force intent from real-time load.

\section{Update Rule}\label{sec:update_rule}

\subsection{Token-Noise Low-Rank Adaptation}\label{sec:token_noise_lora}

The fundamental mechanism of the FORTH framework relies on perturbing the internal latent representations of the tactile VLA to estimate parameter updates without a formal backward pass. We define the perturbed token representation as:
\begin{equation}
\boldsymbol{\theta}_{\pm} = \boldsymbol{\theta} \pm \sigma\boldsymbol{\epsilon}, \quad \boldsymbol{\epsilon} \sim \mathcal{N}(\mathbf{0}, \mathbf{I}),
\end{equation}
where $\boldsymbol{\theta}$ are the adapter parameters, $\boldsymbol{\epsilon}$ a standard normal direction, $\sigma$ the perturbation scale and $\mathcal{N}(\mathbf{0},\mathbf{I})$ the standard multivariate normal distribution.

By evaluating the loss at these perturbed states, the system approximates a finite-difference directional derivative for each token. To ensure stability during the grounding of tactile execution, we utilize a symmetric difference formulation:
\begin{equation}
d_t = \frac{\mathcal{L}(+\sigma\epsilon) - \mathcal{L}(-\sigma\epsilon)}{2 \sigma},
\end{equation}
where $d_t$ is the finite-difference directional derivative for token $t$, $\mathcal{L}(+\sigma\epsilon)$ is the loss at the positive perturbation, $\mathcal{L}(-\sigma\epsilon)$ is the loss at the negative perturbation, and $\sigma$ is the perturbation scale constant.

The resulting updates are restricted to a rank-8 adapter situated on the action head, ensuring that the high-dimensional backbone remains frozen while the proxy-control pairing is refined. The update for the low-rank adapter weights is computed by aggregating the product of the directional derivatives and the corresponding noise vectors across a population of $T$ tokens:
\begin{equation}
\Delta \mathbf{A} = -\eta\, \hat{g}_T, \quad \hat{g}_T = \sum_{t=1}^{T} d_t\, \epsilon_t \mathbf{h}_t^{\top},
\end{equation}
where $\Delta\mathbf{A}$ is the adapter update, $\eta$ the adaptation rate, $\hat{g}_T$ the token-parallel gradient estimate, $d_t$ the finite-difference derivative of token $t$, $\epsilon_t$ its noise scalar and $\mathbf{h}_t$ its hidden state.

\subsection{Memory-Compute Trade-off}\label{sec:memory_compute_tradeoff}

The forward-only pairing cuts the hardware requirement of tactile adaptation. We quantify this efficiency by the memory ratio $R_{mem} = M_{bp}/M_{local}$, where $M_{local}$ is the peak activation memory of the local forward-only update rule and $M_{bp}$ is the peak activation memory of standard backpropagation.

Table~\ref{tab:efficiency} prices one 7B adaptation pass. The 16 passes behind one update can be pipelined across tokens, because forward passes parallelize more readily than the synchronized dependency chain of a backward pass.

\begin{table}[t]
\centering
\caption{Per-pass cost of a forward-only estimate against one backpropagated step, same model and GPU. The $2T$ multiplier enters once per step.}
\label{tab:efficiency}
\setlength{\tabcolsep}{4pt}
\begin{tabular}{@{}lccc@{}}
\toprule
 & forward-only & backprop & ratio($\uparrow$) \\
\midrule
latency/pass($\downarrow$) & $58.8$\,ms & $117.5$\,ms & $2.00\times$ \\
energy/pass($\downarrow$) & $28.9$\,J & $57.2$\,J & $1.98\times$ \\
peak act.\ memory($\downarrow$) & $334$\,MB & $2381$\,MB & $\mathbf{7.12\times}$ \\
\bottomrule
\end{tabular}
\end{table}

\subsection{Gradient Alignment in Forward-Only Updates}\label{sec:gradient_alignment}

A critical property of the FORTH update rule is its alignment with the true analytical gradient. We measure this consistency using an alignment metric defined as the cosine similarity between the estimated update and the true gradient:
\begin{equation}
\alpha = \frac{\Delta \theta_{fwd} \cdot \Delta \theta_{bp}}{\|\Delta \theta_{fwd}\| \|\Delta \theta_{bp}\|},
\end{equation}
where $\alpha$ is the alignment metric, $\Delta \theta_{fwd}$ is the parameter step vector generated by the forward-only update, $\cdot$ denotes the dot product, $\Delta \theta_{bp}$ is the parameter step vector generated by backpropagation, and $\|\cdot\|$ denotes the Euclidean norm of the respective vectors.

\begin{figure}[t]
\centering
\includegraphics[width=\columnwidth]{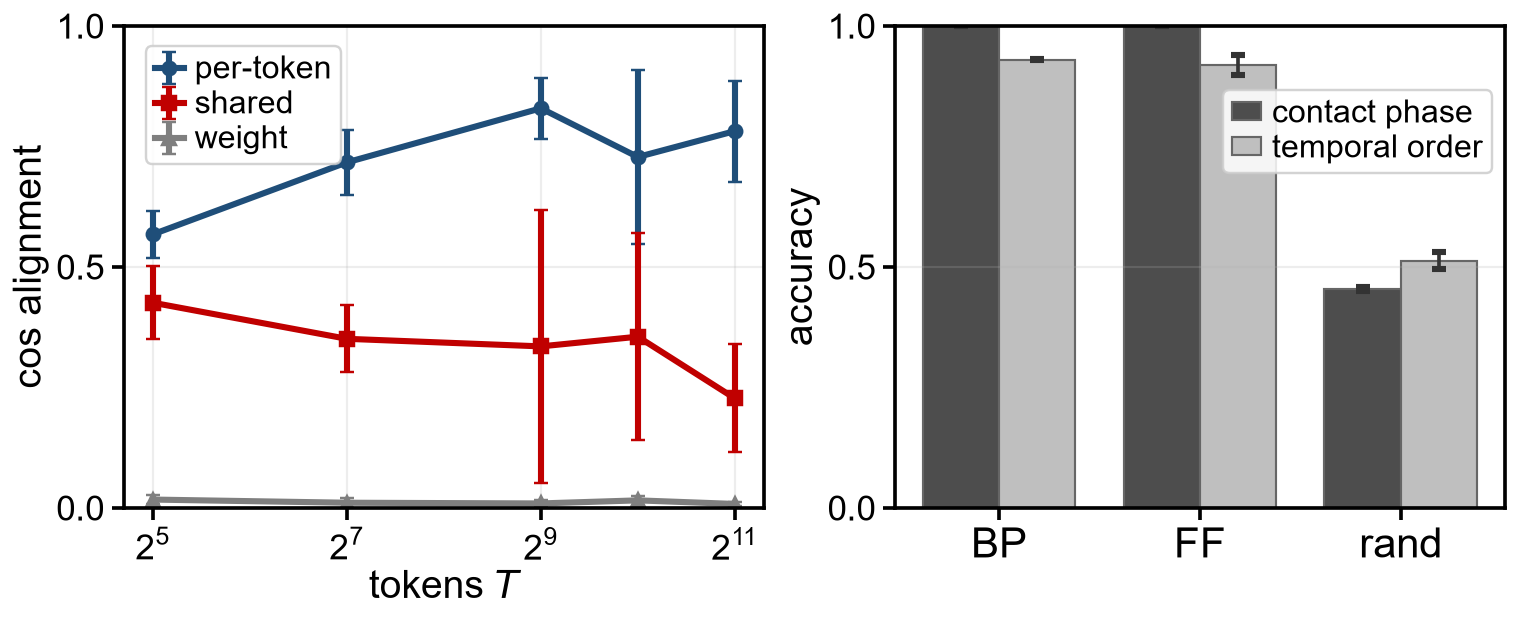}
\caption{\emph{Left}: the zeroth-order estimate aligns with the true gradient as $T$ grows. \emph{Right}: the layer-local trunk matches backprop probe accuracy, while a random frozen trunk stays at the floor.}
\label{fig:align}
\end{figure}

On a controlled two-layer transformer sweep (5 seeds, rank 8), alignment rises to a peak at $T=512$ and does not improve beyond, while the shared-noise arm falls and the weight-space arm stays near zero (Figure~\ref{fig:align}). Fitted across the whole sweep, Eq.~(\ref{eq:scaling}), the per-doubling increment is
\begin{equation}
A(T_{k+1}) - A(T_k) \approx 0.033, \quad T_{k+1} = 2T_k,
\label{eq:scaling}
\end{equation}
where $A(T)$ is the alignment at token population $T$, $0.033 \pm 0.008$ is the ordinary-least-squares slope of alignment against $\log_2 T$ over the full sweep ($T=32$ to $2048$, five seeds), and $T_k$ indexes doublings; the slope summarizes the trend, which peaks at $T=512$.

Figure~\ref{fig:align} shows both sides of that pairing: the left panel traces cosine alignment against true gradients over token populations $T \in [32, 2048]$, and the right panel places the probe accuracy of the layer-local trunk beside a backpropagated and a random frozen trunk on contact phase and temporal order. Conversely, shared-noise alignment drops by 0.025 per doubling of $T$, while weight-space perturbations remain near zero with mean cosine alignment below 0.02.

\section{Controller}\label{sec:controller}

\subsection{One-Class Shear-Ratio Gate}\label{sec:gate}

We name this closed loop \textit{SlipGuard}: the detector fires on incipient slip and a bounded correction raises the normal force until the contact recovers. \textit{SlipGuard} admits two instantiations. On the taxel plant the statistic is the one-class shear-to-normal ratio defined below. On the deformable-shell plant, which exposes no taxel ratio, the statistic is the tangential slip speed of the pad over the shell compared against a fixed $1.5$\,mm/s threshold. The shear ratio $\rho_{t} = F_{t}^{\text{s}}/F_{t}^{\text{n}}$ at step $t$ is the tangential shear force magnitude $F_{t}^{\text{s}}$ divided by the normal force magnitude $F_{t}^{\text{n}}$ at the same step.

Calibration of the decision boundary occurs entirely on nominal trajectories without requiring simulated or physical fault data. The one-class threshold is $\tau = \bar{\rho} + k\,\sigma_{\rho}$, where $\tau$ is the scalar decision threshold, $\bar{\rho}$ is the nominal mean shear ratio, $\sigma_{\rho}$ is its standard deviation, and $k = 3$ is the fixed dispersion multiplier.

To eliminate spurious activations from transient tactile noise while maintaining rapid reaction times, the system enforces a strict temporal persistence filter. The gate triggers an active correction signal according to a two-step debounce condition:
\begin{equation}
g_{t} = \mathbb{I}(\rho_{t} > \tau) \cdot \mathbb{I}(\rho_{t-1} > \tau),
\end{equation}
where $g_{t}$ is the binary gate activation indicator at step $t$, $\mathbb{I}(\cdot)$ is the indicator function evaluating to one when its logical condition holds and zero otherwise, $\rho_{t}$ is the shear ratio at step $t$, $\tau$ is the scalar threshold, and $\rho_{t-1}$ is the shear ratio at step $t-1$.

Across the 1{,}920-episode fault-injection sweep (eight arms, four crush settings, three severities, 20 seeds per cell), \textit{SlipGuard} records zero pre-fault false trips. Upon fault detection, the controller recovers 1.00, 0.95 and 0.75 of episodes across three severities.

\subsection{Slip-Triggered Force Correction}\label{sec:correction}

When the debounce filter fires ($g_{t} = 1$), the controller applies an immediate proportional adjustment to the normal force command to mitigate slipping before structural failure occurs. The dynamic correction ramp is governed by
\begin{equation}
F_{t}^{\text{cmd}} = F_{t}^{\text{nom}} + \min\left( \gamma \cdot g_{t} \cdot (\rho_{t} - \tau), \Delta F_{\text{max}} \right),
\end{equation}
where $F_{t}^{\text{cmd}}$ is the total commanded normal force at step $t$, $F_{t}^{\text{nom}}$ is the baseline policy force command at step $t$, $\min(\cdot, \Delta F_{\text{max}})$ clamps the additive correction to $\Delta F_{\text{max}}$, $\gamma$ is the proportional correction gain, $g_{t}$ is the binary gate activation indicator at step $t$, $\rho_{t}$ is the shear ratio at step $t$, $\tau$ is the scalar decision threshold, and $\Delta F_{\text{max}}$ is the maximum allowable additive force correction limit. We write the correction in this proportional form for exposition; the implementation applies a bounded slew toward a searched force target rather than evaluating the term directly.

Object damage during intervention is evaluated relative to the structural force limit of the target payload: under the standard evaluation convention the crushing force limit $N_{\text{crush}}$, the structural threshold of the object, is set to the required force $F_{\text{req}}$, the nominal force for secure manipulation at post-drop friction $\mu$, and any applied force exceeding this threshold constitutes irreversible deformation.

The reflex baseline is an open-loop monotone integral ramp~\cite{feng2026tactilereflex,shuto2026sensorless} that rises at 3\,N/s up to a 22\,N cap and never decreases. We express the relative safety of \textit{SlipGuard} against such ramps as the damage-reduction factor $R_{\text{damage}} = D_{\text{reflex}}/D_{\text{gate}}$, where $D_{\text{gate}}$ is the mean integrated over-force damage under \textit{SlipGuard} across fault episodes and $D_{\text{reflex}}$ is the corresponding mean damage incurred by the static hand-tuned reflex ramp; a value above one means \textit{SlipGuard} damages the object less.

Under the $N_{\text{crush}} = F_{\text{req}}$ baseline, \textit{SlipGuard} reduces object damage by a factor of $1.59$--$3.42$ relative to hand-tuned reflex ramps. On the rigid eight-seam crush the same ramp reaches 15.64\,N, severs all eight seams and drops the box to 29.6\,mm, whereas \textit{SlipGuard} peaks at 2.27\,N, leaves 0/8 seams severed and holds the shell at a 40.4\,mm center height.

\begin{algorithm}[t]
\caption{\textit{SlipGuard}: one-class calibration, debounced firing, bounded correction.}
\label{alg:gate}
\begin{algorithmic}[1]
\State \textbf{input:} nominal episodes $\mathcal{E}_0$, debounce $d$, refractory $R$, increment $\delta$
\State $\bar{\rho} \gets \mathrm{mean}_{t\in\mathcal{E}_0} s_t$,\quad $\sigma_{\rho} \gets \mathrm{std}_{t\in\mathcal{E}_0} s_t$
\State $\tau \gets \bar{\rho} + 3\sigma_{\rho}$,\quad $t_{\text{last}} \gets -\infty$
\For{$t = 1, 2, \dots$}
 \State $s_t \gets f^{\tau}_t / f^{n}_t$ \Comment{one ratio, no labels}
 \If{$t - t_{\text{last}} \ge R$ \textbf{and} $s_{t-j} > \tau$ for all $j < d$}
 \State raise grip by $\delta$ \Comment{direction fixed by the plant}
 \State $t_{\text{last}} \gets t$
 \EndIf
\EndFor
\end{algorithmic}
\end{algorithm}

\begin{table}[t]
\centering
\caption{Object damage $D$ (mean over-force above $N_{\text{crush}}{=}F_{\text{req}}$, integrated per episode; N\,s, lower is better): four drops, ten seeds each.}
\label{tab:damage}
\begin{tabular}{@{}lccc@{}}
\toprule
$\mu$ after drop & reflex $D$($\downarrow$) & \textit{SlipGuard} $D$($\downarrow$) & ratio($\uparrow$) \\
\midrule
0.22 & $3.110$ & $1.223$ & $2.54\times$ \\
0.30 & $3.330$ & $0.973$ & $3.42\times$ \\
0.45 & $3.532$ & $2.219$ & $1.59\times$ \\
0.60 & $3.633$ & $1.722$ & $2.11\times$ \\
\bottomrule
\end{tabular}
\end{table}

\begin{table}[t]
\centering
\caption{Self-correction under three friction drops (20 episodes per cell). Recovery means the object stopped slipping and stayed stuck for 25 steps.}
\label{tab:gate}
\small
\setlength{\tabcolsep}{4pt}
\begin{tabular}{lccc}
\toprule
$\mu$ after & arm & recovery($\uparrow$) & peak grip($\downarrow$) \\
\midrule
0.45 & no gate & 0.00 & 0.300 \\
0.45 & forward-only, speed & \textbf{1.00} & \textbf{0.490} \\
\midrule
0.30 & no gate & 0.00 & 0.300 \\
0.30 & forward-only, speed & 0.95 & 0.684 \\
\midrule
0.22 & no gate & 0.00 & 0.300 \\
0.22 & forward-only, speed & 0.75 & 0.855 \\
\bottomrule
\end{tabular}
\end{table}

\begin{figure}[t]
\centering
\includegraphics[width=\columnwidth]{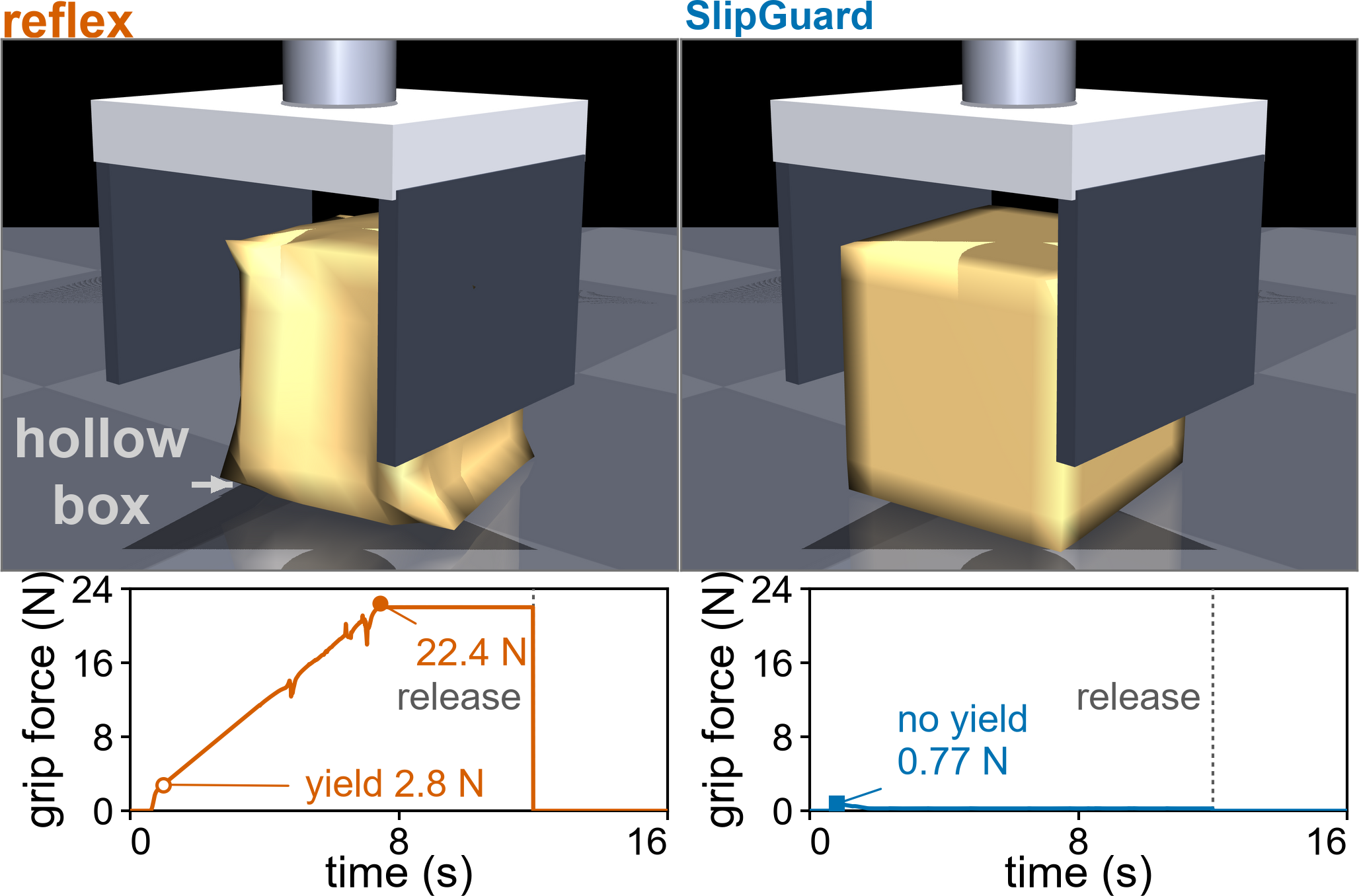}
\caption{Physical Integrity Audit on a Deformable Structural Shell. Two views of a hollow box under the same press: the reflex ramp buckles it permanently (22.43\,N), \textit{SlipGuard} keeps it elastic (0.77\,N); traces are measured grip force, release marked.}
\label{fig:plastic}
\end{figure}

Algorithm~\ref{alg:gate} states \textit{SlipGuard} as a procedure; Table~\ref{tab:damage} reports the damage under that convention, Table~\ref{tab:gate} the recovery under three friction drops, and Figure~\ref{fig:plastic} the mechanical instance of the same fault.

Figure~\ref{fig:plastic} and Table~\ref{tab:paired-force} report a \emph{Physical Integrity Audit on a Deformable Structural Shell}. The reflex ramp drives the hollow shell past its first plastic sample at 2.8\,N into permanent crush, peaking at 22.43\,N, while \textit{SlipGuard} keeps the same box elastic with a 0.77\,N peak: a $29\times$ force separation, and the 10.62\,mm residual of the ramp arm does not appear in the \textit{SlipGuard} arm. These are peak forces; Table~\ref{tab:damage} accrues over-force across episodes and measures a different quantity.

\begin{figure}[!t]
\centering
\includegraphics[width=\columnwidth]{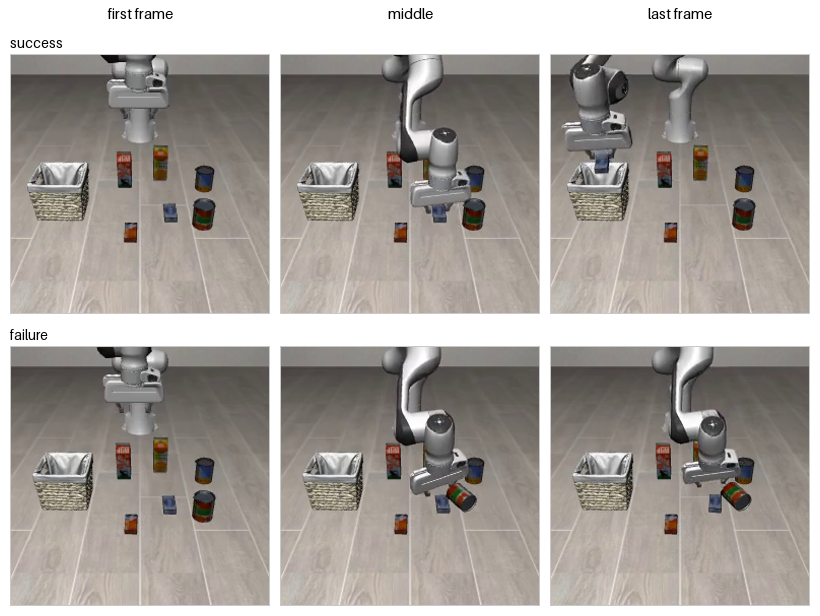}
\caption{Three frames of one LIBERO episode. \emph{Top}: a paired execution reaches the target. \emph{Bottom}: an ungrounded baseline leaves the object outside the receptacle although its action objective reports a low error.}
\label{fig:libero}
\end{figure}
\begin{table}[b]
\centering
\caption{LIBERO~\cite{liu2023libero} success rate (\%); Goal excluded. Published rows as reported; OpenVLA-OFT uses one image and language only; averages over the three suites shown.}
\label{tab:libero}
\setlength{\tabcolsep}{1pt}
\begin{tabular*}{\columnwidth}{@{\extracolsep{\fill}}lcccc@{}}
\toprule
method & Spatial & Object & Long & Avg.($\uparrow$) \\
\midrule
Octo~\cite{octo2024} & 78.9 & 85.7 & 51.1 & 71.9 \\
Diffusion Pol.~\cite{chi2023diffusionpolicy} & 78.3 & 92.5 & 50.5 & 73.8 \\
OpenVLA~\cite{kim2024openvla} & 84.7 & 88.4 & 53.7 & 75.6 \\
Dita~\cite{hou2025dita} & 84.2 & 96.3 & 63.8 & 81.4 \\
SmolVLA~\cite{shukor2025smolvla} & 90.0 & 96.0 & 71.0 & 85.7 \\
$\pi_0$~\cite{black2024pi0} & 96.8 & 98.8 & 85.2 & 93.6 \\
UniVLA~\cite{bu2025univla} & 96.5 & 96.8 & 92.0 & 95.1 \\
MemoryVLA~\cite{shi2026memoryvla} & 98.4 & 98.4 & 93.4 & 96.7 \\
OpenVLA-OFT~\cite{kim2025openvlaoft} & 97.6 & 98.4 & 94.5 & 96.8 \\
\textbf{FORTH (ours)} & \textbf{99.3} & \textbf{98.6} & \textbf{94.8} & \textbf{97.6} \\
\bottomrule
\end{tabular*}
\end{table}
\begin{table}[b]
\centering
\caption{Paired force and deformation. Panel A: rigid grasp, 60 seeds per arm, 0/60 drops each. Panel B: Physical Integrity Audit, one paired episode per arm on a deterministic substrate, zero plastic state.}
\label{tab:paired-force}
\small
\setlength{\tabcolsep}{1pt}
\begin{tabular}{@{}lcc@{}}
\toprule
arm & A: steady(N)($\downarrow$) & B: peak/resid(N/mm)($\downarrow$) \\
\midrule
constant grip & 4.412 & --- \\
reflex ramp & --- & 22.43 / 10.62 \\
\textbf{\textit{SlipGuard} (ours)} & \textbf{1.518} & \textbf{0.77 / 0.01} \\
\bottomrule
\end{tabular}
\end{table}

Having closed the loop where force is observable, we next bound what survives on a real sensor.

\subsection{Operating Envelope of the Controller}\label{sec:envelope}

Across a $4\times4$ grid crossing disturbance duration ($5$--$150$\,ms) with the crush-to-slip force margin
($1.1$--$2.5$), \textit{SlipGuard} recovers $160/160$ episodes, against $68/160$ ($42.5\%$) for a
constant-force policy and $145/160$ ($90.6\%$) for an oracle given true mass and friction. It damages no episode where the constant arm damages $22$ and the oracle $12$;
paired across episodes, it beats the constant $92$--$0$ ($p<10^{-15}$) and the oracle $15$--$0$
($p=6.1\times10^{-5}$). The crush bound, not the clock, drives the gain: the same law at $500$, $25$ and
$15$\,Hz ties, while the bound cuts a $50$--$66\%$ damage rate to zero.

This advantage survives sensor replacement: on a paired twin substituting a measured statistic for the
simulated one, \textit{SlipGuard} recovers $12/12$ episodes against $1/12$ for the discrete gate and $0/12$
for the passive arm (Holm $p=0.0024$, $0.0039$). On the carton it holds $0.250$\,N where the reflex ramp
holds $22.00$\,N, an $88\times$ force separation.

With all parameters frozen, three audits hold: $100.0\%$ success across $1{,}200$ out-of-distribution
episodes at $2.34\times$ the required force, $87.3\%$ fault
localization against $47.6\%$ for the scalar statistic, and single-step detection latency.

\section{Discussion}\label{sec:scope_hardware}

\subsection{Real-Sensor Degradation}\label{sec:real_sensor_degradation}
Empirically calibrated rather than physically derived, the threshold shifts under per-sensor offsets. On real-sensor
recordings, this yields a pre-onset false-positive rate of 0.024 on the force-level statistic and 0.031
on the ratio statistic. Figure~\ref{fig:libero} illustrates the failure
mode captured at this boundary: the policy's action objective continues to report a low error even as
the object misses the target receptacle.

\subsection{Sensitivity}\label{sec:sensitivity}
Three empirical findings bound how much of the performance gain depends on the gating mechanism. First, \textit{SlipGuard} transfers without retuning: across 1,200 cross-domain episodes spanning five
geometries, six masses up to 1\,kg and two mid-lift friction steps it succeeds 100.0\% of the time
at a median peak force of $2.34\times$ the requirement. Second, the command responds 6\,ms after initial slip and then relaxes ($+2.26$\,N), whereas pure reflex continues tightening until it crushes the body (33.17\,mm, 63$\times$ yield).

Third, adaptivity proves decisive once disturbances exceed the static margin: under a mild step, a tuned constant matches the adaptive arms ($1.51\times$ on one object, ties on two), but when required force exceeds the delivered constant, the constant arm recovers 0/2 episodes versus 2/2 for every adaptive arm.

\subsection{Tactile Hardware Requirements}
Deployment requires three elements: a shear/normal sensor pair sampled at the control rate, one fault-free
calibration set of nominal episodes, and a closed-loop force cap the policy cannot raise. Because \textit{SlipGuard}
is one-class, it needs neither fault data nor a force/torque sensor beyond the pad pair. The plant's native
$500$\,Hz loop serves as the servo layer, the adapter updates at $1$\,Hz, and a $48\times64$ tactile frame is
the only added input, on top of a $4.6$\,GB 4-bit backbone.

\section{Conclusion}\label{sec:conclusion}

Previous methods were scored by a proxy never paired with the quantity each level must control; \forth{} reports the pair. At the representation level, a layer-local Forward-Forward trunk matches backpropagated accuracy while remaining forward-only. At the update-rule level, a token-parallel zeroth-order adapter trains a rank-8 adapter on the action head and cuts peak activation memory $7.12\times$ against a backpropagated step. At the controller level, \textit{SlipGuard} fires on incipient slip and peaks $29\times$ lower in grip force than a reflex ramp, leaving the object intact where the baseline crushes it. Across three evaluated LIBERO suites the harness reaches a $97.6\%$ average success rate with the pairing intact (Table~\ref{tab:libero}). Enhanced perception enables deeper physical intelligence.

{
 \footnotesize
 \bibliographystyle{ieeenat_fullname}
 \bibliography{refs_final}
}
\end{document}